\documentclass[a4paper,fleqn]{cas-dc}

\usepackage{lineno,hyperref}
\modulolinenumbers[5]

\usepackage[numbers]{natbib}

\usepackage{amssymb}
\usepackage{pifont}
\usepackage{amsfonts}
\usepackage{amsmath}
\usepackage{algorithm}
\usepackage{algpseudocode}
\usepackage{graphicx,amssymb,amstext}
\usepackage{bm,bbm}
\usepackage[caption=false,labelfont=footnotesize,singlelinecheck=true]{subfig}
\usepackage{multirow}
\usepackage{amsmath,algorithm,tabularx}
\usepackage[inline,shortlabels]{enumitem}
\usepackage{url}

\def\tsc#1{\csdef{#1}{\textsc{\lowercase{#1}}\xspace}}
\tsc{WGM}
\tsc{QE}
\tsc{EP}
\tsc{PMS}
\tsc{BEC}
\tsc{DE}

\makeatletter
\newcommand{\multiline}[1]{%
  \begin{tabularx}{\dimexpr\linewidth-\ALG@thistlm}[t]{@{}X@{}}
    #1
  \end{tabularx}
}
\makeatother

\begin{document}
\let\WriteBookmarks\relax
\def\floatpagepagefraction{1}
\def\textpagefraction{.001}
\shorttitle{Ozone Prediction Using the Lasso}
\shortauthors{J. Lv \& X. Xu}

\title [mode = title]{Prediction of daily maximum ozone levels using Lasso sparse modeling method}                      

\author{Jiaqing Lv}[orcid=0000-0002-7992-6716]
\ead{lvjiaqing@gmail.com}

\author{Xiaohong Xu}
\ead{xxu@uwindsor.ca}

\address{Civil \& Environmental Engineering, University of Windsor, Canada}

\begin{abstract}
This paper applies modern statistical methods in prediction of the next-day maximum ozone concentration, as well as the maximum 8-hour-mean ozone concentration of the next day. The model uses a large number of candidate features, including the present day's hourly concentration level of various pollutants, as well as the meteorological variables of the present day's observation and the future day's forecast values. In order to solve such an ultra-high dimensional problem, the least absolute shrinkage and selection operator (Lasso) was applied. The $L_1$ nature of this methodology enables the automatic feature dimension reduction, and a resultant sparse model. The model trained by 3-years data demonstrates relatively good prediction accuracy, with RMSE= 5.63 ppb, MAE= 4.42 ppb for predicting the next-day's maximum $O_3$ concentration, and RMSE= 5.68 ppb, MAE= 4.52 ppb for predicting the next-day's maximum 8-hour-mean $O_3$ concentration. Our modeling approach is also compared with several other methods recently applied in the field, and demonstrates superiority in the prediction accuracy.

\end{abstract}


\maketitle

\section{Introduction}

Ground-level ozone is one of the most important pollutants whose adverse effects on human health have been intensively studied \cite{goldsmith1969experimental,saez2001comparing}. Further, its impact on agriculture has been estimated to cause several billion dollars of annual economic loss from its negative impact on plants and crops \cite{larsen1991air}. As a consequence, accurate forecast of ground-level ozone concentration plays an important role in air quality management, environmental monitoring and control.

The concentration of the tropospheric $O_3$ may have a profound interaction with other pollutants such as nitrogen oxides ($NO_x$) etc through complex photochemical reactions \cite{barrero2006prediction}. Also, meteorological factors such as temperature, humidity, solar radiation, and wind speed \cite{barrero2006prediction} often have an impact on $O_3$ levels. Many environmental agencies rely on ozone concentration forecasting framework for decision making. With the advent of the big data age, machine learning and regression methods have been more and more applied in the accurate modeling and forecast of ozone concentration. Generally speaking, within the ozone prediction problem, there are mainly two types of machine learning approaches: statistical modeling and non-statistical methodologies. Methods such as time-series inference (e.g., ARMA or ARIMA forecasting) \cite{robeson1990evaluation}, multiple linear regression \cite{barrero2006prediction,ghazali2010transformation,sousa2006prediction}, ridge regression \cite{salazar2008development} belong to the formal category, while the non-statistical methods include ANN \cite{gomez2006neural,abdul2002assessment} or other neural network type of methods \cite{dutot200724,coman2008hourly,prybutok2000comparison,wang2003prediction,yi1996neural}, support vector regression \cite{hajek2012ozone,lu2008ground} and others \cite{sokhi2006prediction}.

Among the ozone prediction literature, many aim to target the next-day's maximum concentration \cite{dutot200724,wang2003prediction,yi1996neural,robeson1990evaluation}. On the other hand, the next-day's maximum of the 8-hour mean concentration \cite{link3} also plays a significant role in the environmental regulations/ control in Canada. Therefore we consider both of these two variables to be the target variable we aim to predict in our research.

The aforementioned statistical and non-statistical methods in the ozone prediction literature usually are conducted on a carefully-chosen set of features. Quite often the number of variables in the model is less than ten (10). On the other hand, due to the renowned ``curse of dimensionality'' phenomenon \cite{greblicki2008nonparametric,buhlmann2011statistics} in the machine learning and artificial intelligence literature, it is impossible to conduct large dimensional modeling without proper feature selection as a first step before the modeling. However, the classical feature selection techniques such as AIC, BIC, forward selection, backward selection all have drawbacks \cite{efron2016computer}. On the other hand, as the least absolute shrinkage and selection operator (Lasso) has been invented in the late 1990s \cite{tibshirani1996regression}, its $L_1$ design nature allows this approach to automatically reduce the redundant features and render a sparse model. Since then, the Lasso has become a popular approach in modern-day high-dimensional statistics \cite{buhlmann2011statistics}. It has been applied in biometrics \cite{mikula2009effects}, power systems \cite{lv2012prediction,mo2020power}, energy \cite{lv2020very}, etc. So far as we know, it has not been applied in pollutant modeling areas yet.

The rest of the paper will be organized as follows. Section \ref{sbsec:Data} will discuss the data we are using in this research. Section \ref{sbsec:Lasso} will give a description about the LASSO methodology and the algorithm to solve Lasso.  Section \ref{sec:Results} will show the modeling result by our method. The prediction performance of some competing methods including ARMA model, support vector regression, multiple linear regression, and ridge regression will be shown as a comparison to our method. The discussions and future work will be shown in Section \ref{sec:Discussions}.

The contributions of this paper include:
\begin{enumerate}
  \item Use sparse machine learning algorithms (Lasso) to model and predict the next-day maximum ozone concentration, as well as the next-day maximum 8-hour-mean ozone concentration. The $L_1$ design enables the automatic feature selection from more than hundreds of dimensions of candidate features. In our research we use a large number of features available from the Environmental Canada official website. However, the same technique can also apply if some of these features were not available to the user.
  \item We compare our modeling methods with several other competing methods in the field. Namely, they are multiple linear regression (MLR), ridge regression, ARMA modeling, and support vector regression. Our Lasso shows superior prediction accuracy compared to the other approaches on the same set of features.
\end{enumerate}

\section{Materials and Methods}

\subsection{Data Collection and Pre-processing}\label{sbsec:Data}
\begin{figure}
\centering
\includegraphics[width=85mm]{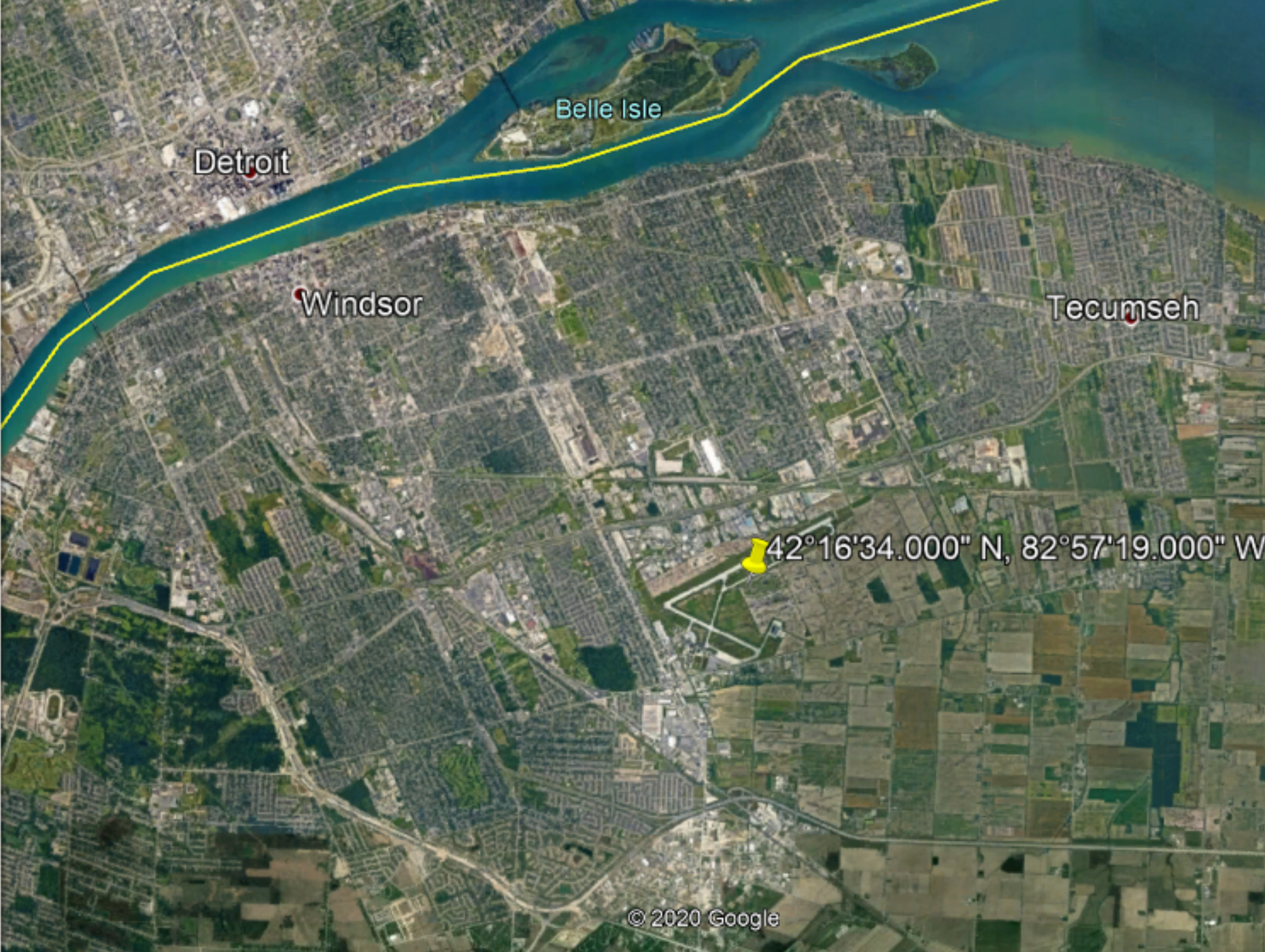}
 \caption{The illustration of the map of Windsor and the location of the air quality and meteorological monitoring station (N $42^\circ$$16'$$34''$  W $82^\circ$$57'$$19''$)}\label{fig:Map}
\end{figure}

The study site is the city of Windsor (N $42^\circ$$16'$$34''$  W $82^\circ$$57'$$19''$), Ontario, Canada; the pollutant concentration observations and the meteorological data from 2014 to 2017 were downloaded downloaded from the Ontario Ministry of The Environment, Conservation and Parks \cite{link1} and Environment Canada \cite{link2}. The map of the city of Windsor and the location of the air quality and meteorological monitoring station is illustrated in Fig. \ref{fig:Map}.

The candidate features we used in our modeling are shown in Table \ref{tab:Features}. In our modeling studies, we not only use all of them, but also include the interactive features among them. The detailed modeling will be shown in Section \ref{sec:Results}.

\begin{table}[width=.9\linewidth,cols=4,pos=h]
\caption{Air quality and meteorological features used in our modeling}\label{tab:Features}
\begin{tabular*}{\tblwidth}{@{} LL@{} }
\toprule
\normalsize{Model Output} & \scriptsize{Next-day's maximum $O_3$ (ppb)} \\
~& \scriptsize{Next-day's maximum 8-hour-mean $O_3$ (ppb)}  \\
\midrule
\normalsize{Model Input} & \scriptsize{Present-day's hourly $O_3$ (ppb)}  \\
~ & \scriptsize{Present-day's hourly $SO_2$ (ppb)}  \\
~ & \scriptsize{Present-day's hourly $NO$ (ppb)}  \\
~ & \scriptsize{Present-day's hourly $NO_2$ (ppb)}  \\
~ & \scriptsize{Present-day's hourly $NO_X$ (ppb)}  \\
~ & \scriptsize{Present-day's hourly $CO$ (ppb)}  \\
~ & \scriptsize{Present-day's hourly $PM_{2.5}$ ($mg/m^3$)}  \\
~ & \scriptsize{Present-day's max/min/mean $O_3$ (ppb)}  \\
~ & \scriptsize{Present-day's max/min/mean $SO_2$ (ppb)}  \\
~ & \scriptsize{Present-day's max/min/mean $NO$ (ppb)}  \\
~ & \scriptsize{Present-day's max/min/mean $NO_2$ (ppb)}  \\
~ & \scriptsize{Present-day's max/min/mean $NO_X$ (ppb)}  \\
~ & \scriptsize{Present-day's max/min/mean $CO$ (ppb)}  \\
~ & \scriptsize{Present-day's max/min/mean $PM_{2.5}$ ($mg/m^3$)}  \\
~ & \scriptsize{Present-day's max/min/mean $O_3$ (ppb)}  \\
~ & \scriptsize{Present-day's hourly temperature (${}^\circ C$)}  \\
~ & \scriptsize{Present-day's hourly dew point (${}^\circ C$)}  \\
~ & \scriptsize{Present-day's hourly real humidity (\%)}  \\
~ & \scriptsize{Present-day's hourly wind direction (deg)}  \\
~ & \scriptsize{Present-day's hourly wind speed (km/h)}  \\
~ & \scriptsize{Present-day's hourly visibility (km)}  \\
~ & \scriptsize{Present-day's hourly atmospheric pressure (kPa)}  \\
~ & \scriptsize{Present-day's max/min/mean temperature (${}^\circ C$)}  \\
~ & \scriptsize{Present-day's max/min/mean dew point (${}^\circ C$)}  \\
~ & \scriptsize{Present-day's max/min/mean rel humidity (\%)}  \\
~ & \scriptsize{Present-day's max/min/mean wind direction (deg)}  \\
~ & \scriptsize{Present-day's max/min/mean wind speed (km/h)}  \\
~ & \scriptsize{Present-day's max/min/mean visibility (km)}  \\
~ & \scriptsize{Present-day's max/min/mean atmospheric pressure (kPa)}  \\
~ & \scriptsize{Present-day's hourly temperature (${}^\circ C$)}  \\
~ & \scriptsize{Next-day's hourly dew point (${}^\circ C$)}  \\
~ & \scriptsize{Next-day's hourly rel humidity (\%)}  \\
~ & \scriptsize{Next-day's hourly wind direction (deg)}  \\
~ & \scriptsize{Next-day's hourly wind speed (km/h)}  \\
~ & \scriptsize{Next-day's hourly visibility (km)}  \\
~ & \scriptsize{Next-day's hourly atmospheric pressure (kPa)}  \\
~ & \scriptsize{Next-day's max/min/mean temperature (${}^\circ C$)}  \\
~ & \scriptsize{Next-day's max/min/mean dew point (${}^\circ C$)}  \\
~ & \scriptsize{Next-day's max/min/mean rel humidity (\%)}  \\
~ & \scriptsize{Next-day's max/min/mean wind direction (deg)}  \\
~ & \scriptsize{Next-day's max/min/mean wind speed (km/h)}  \\
~ & \scriptsize{Next-day's max/min/mean visibility (km)}  \\
~ & \scriptsize{Next-day's max/min/mean atmospheric pressure (kPa)}  \\
\bottomrule
\end{tabular*}
\end{table}

\subsection{Machine Learning Methods for Sparse Modeling}\label{sbsec:Lasso}

In this section, we give a brief description on the machine learning algorithms we use in the ozone forecast problem. Generally speaking, in system modeling, the researchers are given data of the form
\[
D_n = \{(\bm{X}_1, Y_1), (\bm{X}_2, Y_2), \cdots, (\bm{X}_n, Y_n)\},
\]
where $\bm{X}_i \in \mathbb{R}^p$, $i\in\{1,\cdots,n\}$ is a $p$-dimensional vector denoting the input observations, and $Y_i \in \mathbb{R}^1$, $i\in\{1,\cdots,n\}$ is a 1-dimensional variable denoting the output response. The user is interested to learn the mapping $f: X \to Y$, so that by applying the model $f(\cdot)$ for the future observations $\bm{X}_i^{[new]}$, the predicted response $\hat{f}(\bm{X}_i^{[new]})$ can be as close as possible to the underlying rue response $Y_i^{[new]}$.

In our ground-level ozone concentration modeling problem, we try to build two models. The first model $f_1(\cdot)$ tries to predict the next-day maximum ozone concentration level; in other words, $Y_i$ corresponds to the next-day maximum $O_3$ concentration level. The input variable $\bm{X}_i$ corresponds to all the variables in Table \ref{tab:Features}. In the second model $f_2(\cdot)$, $Y_i$ corresponds to the next-day maximum 8-hour-mean concentration level, while $\bm{X}_i$ includes all those features in Table \ref{tab:Features}, together with the 8-hour-mean concentration values of $O_3$ between 0 am to 4 pm of the present day. It is worth mentioning that according to the definition of the ``8-hour-mean concentration'' \cite{link3}, the value at 4 pm already include the composition of the hourly value at 11 pm of the same day.

Before introducing the sparse learning method used in our research, let's start from a classical regression method called ``multiple linear regression'' (MLR) and a more advanced technique called ridge regression. It is worth mentioning that these two techniques have been applied in the context of air pollutant modeling in \cite{barrero2006prediction,ghazali2010transformation,sousa2006prediction} and in \cite{salazar2008development}. 

There could be various ways for system modeling. If the model has completely unknown structure due to e.g., physical limitations, then it is always reasonable by starting from a linear structure assumption
\begin{equation}
Y_i = \beta_0+\sum_{j=1}^p \beta_j X_{ij}+\varepsilon_i,
\label{eq:LrMd}
\end{equation}
where $\varepsilon_i$ is an innovation process, or the noise process, $\{\beta_0, \cdots, \beta_p\}$ are the coefficients that represent the weight of each feature. To identify the linear model represented by (\ref{eq:LrMd}), it is equivalent to identify only $\{\beta_0, \cdots, \beta_p\}$.

Statistical researchers also often writes the data form of (\ref{eq:LrMd}) in the following design matrix form
\begin{equation} \label{matrixeq}
\bm{Y}=\bm{X} \bm{\beta} +\bm{\varepsilon},
\end{equation}
where the design matrix $\bm{Y}$, $\bm{X}$, $\bm{\beta}$, $\bm{\varepsilon}$ correspond to 
\[
\bm{Y}=\begin{bmatrix}  Y_1  \\ \vdots \\ Y_n \end{bmatrix}, \qquad
\bm{X}=\begin{bmatrix} 1 & X_{11} & \cdots & X_{1p} \\ \vdots & \vdots & \ddots & \vdots \\ 1 & X_{n1} & \cdots & X_{np} \end{bmatrix},
\]
\[
\bm{\beta} = \begin{bmatrix}  \beta_0,  \beta_1, \cdots, \beta_p \end{bmatrix}^{\top}, \qquad
\bm{\varepsilon}=\begin{bmatrix}  \varepsilon_1, \cdots, \varepsilon_n \end{bmatrix}^{\top},
\]
and $X_{ij}$, corresponds to the $j$-th coordinate of the $p$-dimensional observation $\bm{X}_i$, $i=1,\cdots,n$.

The most classical approach to solve the linear model in (\ref{eq:LrMd}) is through the minimization of the mean square error, and this approach is often called ``multiple linear regression'' (MLR)
\begin{equation}
\hat{\bm{\beta}}_{MLR}= \arg\min_{\bm{\beta}} \frac{1}{n} ||\bm{Y}-\bm{X}\bm{\beta}||_2^2,
\end{equation}
where $\frac{1}{n} ||\bm{Y}-\bm{X}\bm{\beta}||_2^2$ is equivalent to $\frac{1}{n} \sum_{i=1}^n (Y_i-\beta_0-\sum_{j=1}^p \beta_j X_{ij})^2 $. The solution of the MLR has the direct analytical form \cite{seber2012linear}
\begin{equation}
\hat{\bm{\beta}}_{MLR} = (\bm{X}^T \bm{X})^{-1} (\bm{X}^T \bm{Y}).
\label{eq:LrRg}
\end{equation}

The multiple linear regression provides a simple but straightforward solution to the linear model. The similar principle, i.e., minimization of the mean square loss, has also been applied in many engineering areas such as signal processing \cite{pawlak2007nonparametric}, image recognition \cite{liao1996image}, etc. However, the MLR naturally processes some drawbacks that inherently makes it infeasible as a modern statistical tool. These inherent deficiencies include
\begin{enumerate}
  \item Numerical instability. When new observations come to be available, the estimated weights $\{\hat{\beta}_j, j=1,\cdots,p\}$ can change dramatically. Even worse, some weights can change from positive to negative, or vice versa. This phenomenon can create undesirable scenarios when the researcher wants to e.g., explain the positive or negative influence of a feature.
  \item The matrix $\bm{X}^T \bm{X}$ may not be full rank. In this way there would be some problems in the matrix inversion process in the solution (\ref{eq:LrRg}).
\end{enumerate}

Targeting these drawbacks, a more modern approach called ``ridge regression'' was developed in the 1970s \cite{hoerl1970ridge,hoerl1970ridge2}. The original intuition was to add a penalty term to the weights, so that the numerical instability problem and the matrix inversion problem can be overcome. The ridge regression tries to find the solution to the following penalized criterion function
\begin{equation}
\hat{\bm{\beta}}_{ridge}= \arg\min_{\bm{\beta}} \Big( \frac{1}{n} ||\bm{Y}-\bm{X}\bm{\beta}||_2^2 +\lambda \sum_{j=1}^p \beta_j^2 \Big),
\label{eq:RidgeCr}
\end{equation}
where $\lambda$ is a regularization parameter that needs to be selected beforehand. This parameter controls the balance between the least square loss and the penalty of the weights. In practice, the user can always use some re-sampling methods to select this parameter, e.g., selecting the value that minimize the cross-validation error.

The ridge regression approach also has close-form solution in the following form,
\begin{equation}
\hat{\bm{\beta}}_{ridge} = (\bm{X}^{\top} \bm{X}+n\lambda \bm{I}_p)^{-1} (\bm{X}^{\top} \bm{Y}),
\label{eq:Ridge}
\end{equation}
where $\bm{I}_p$ denotes the $p \times p$ identity matrix.

The solution in (\ref{eq:Ridge}) ensures that ridge regression successfully overcomes the numerical instability drawback processed by the MLR. As a matter of fact, in many applications, the users observe significant improvement by using the ridge regression compared with the MLR. The ridge regression has recently been more and more often applied in many fields of engineering and science \cite{jain1985ridge,jayasekara2006derivation}.

Following the same line as MLR and ridge regression, another technique was developed in the late 1990s \cite{tibshirani1996regression} and was named ``least absolute shrinkage and selection operator'' (Lasso). Similar to the ridge regression, the Lasso also targets to minimize the mean square loss together with a penalty term. However, the uniqueness of the Lasso lies in that the penalty term is in the form of $L_1$ norm rather than the $L_2$ norm as in the ridge regression case. Specifically, the Lasso approach aims to find the solution according to the following criterion 
\begin{equation}
\hat{\bm{\beta}}_{Lasso}= \arg\min_{\bm{\beta}} \Big( \frac{1}{n} ||\bm{Y}-\bm{X}\bm{\beta}||_2^2 +\lambda \sum_{j=1}^p |\beta_j| \Big),
\label{eq:LassoCr}
\end{equation}
where similar to the ridge regression case, here $\lambda$ is the regularization term which should be specified before the predictive modeling.

According to the optimization theory \cite{boyd2004convex}, the ridge regression in (\ref{eq:RidgeCr}) and the Lasso regression in (\ref{eq:LassoCr}) can be alternatively written in the following equivalent primal forms,
\begin{equation}
\hat{\bm{\beta}}_{ridge,primal}= \arg\min_{\bm{\beta}, ||\bm{\beta}||_2 < s} \Big( \frac{1}{n} ||\bm{Y}-\bm{X}\bm{\beta}||_2^2  \Big),
\label{eq:RidgePrCr}
\end{equation}

\begin{equation}
\hat{\bm{\beta}}_{Lasso,primal}= \arg\min_{\bm{\beta}, ||\bm{\beta}||_1 < s} \Big( \frac{1}{n} ||\bm{Y}-\bm{X}\bm{\beta}||_2^2  \Big),
\label{eq:LassoPrCr}
\end{equation}
where $||\bm{\cdot}||_2$ and $||\bm{\cdot}||_1$ denote the $L_2$ norm and the $L_1$ norm. More specifically, $||\bm{\beta}||_2 = (\sum_{j=1}^p \beta_j^2)^{1/2}$ and $||\bm{\beta}||_1=\sum_{j=1}^p|\beta_j|$. It is worth to mention that there is one-to-one mapping between the parameter $s$ in primal form and the parameter $\lambda$ in the previous dual form. 

\begin{figure}[t]
\centering
\includegraphics[width=70mm]{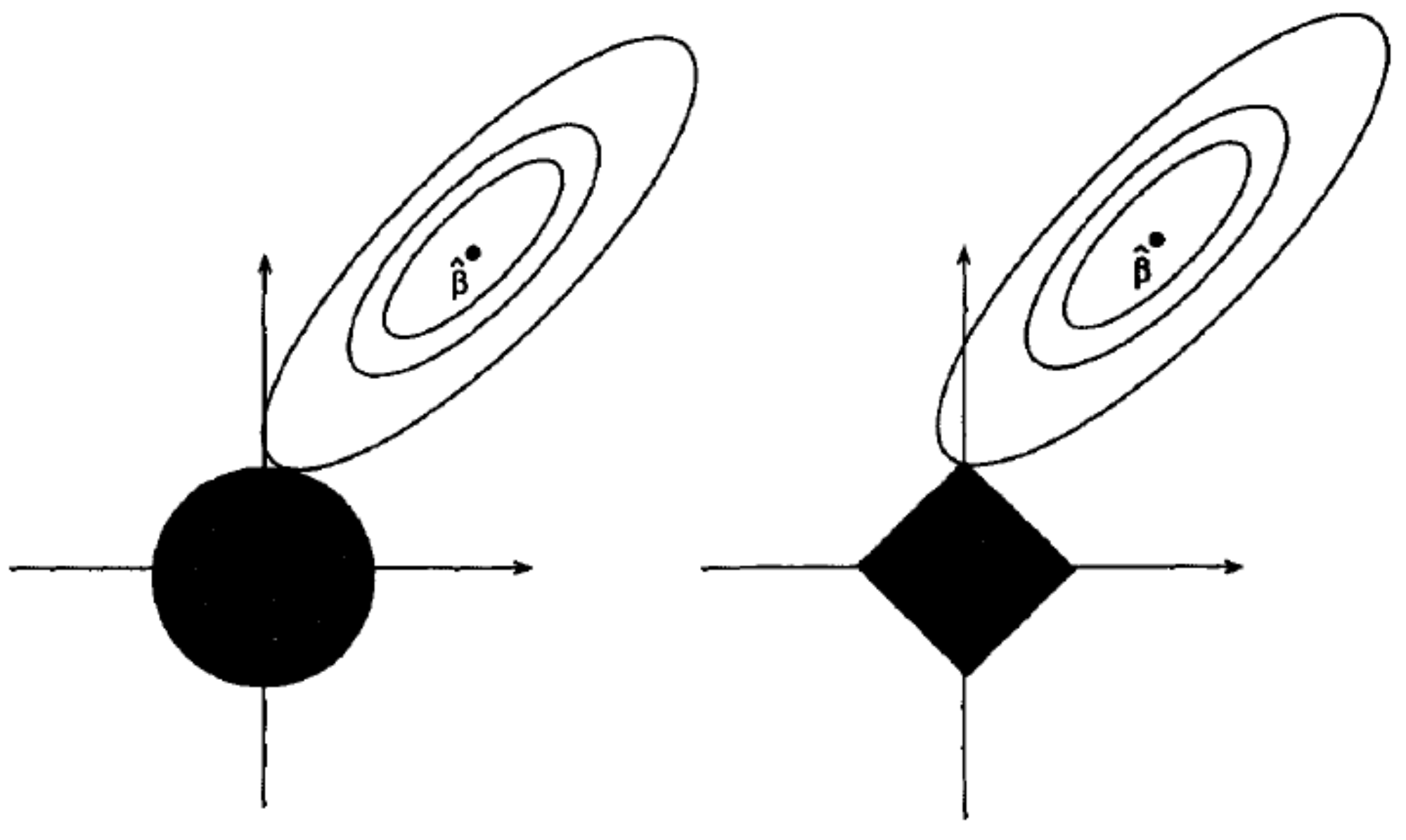}
 \caption{The comparison between ridge regression and the Lasso.}\label{fig:L2L1Compare}
\end{figure}

Equation (\ref{eq:RidgePrCr}) and (\ref{eq:LassoPrCr}) suggest the difference between the Lasso and ridge regression, which is furthermore illustrated in Fig. \ref{fig:L2L1Compare}. 

Fig. \ref{fig:L2L1Compare} shows a 2-dimensional regression case as an example. The elliptical shapes correspond to the same level of mean square error caused by different pairs of coefficients. The shaded area correspond to the restraints of coefficients. The $L_2$ penalty $||\bm{\beta}||_2=(\beta_1^2+\beta_2^2)^{1/2}$ in this case corresponds to the square shape, whereas the $L_1$ penalty $||\bm{\beta}||_1=|\beta_1|+|\beta_2|$ in this case corresponds to the square shape. The pre-specified regularization parameter $\lambda$ (or equivalently $s$) controls how large the shaded zone is. The point where the elliptical shape touches the shaded area corresponds to the solution of the regression given the pre-specified parameter $\lambda$ (or $s$). From Fig. \ref{fig:L2L1Compare}, one can see that there is a chance on the right panel that the solution can ``hit the corner'', meaning that one feature is assigned zero weight and completely eliminated from the regression model in the Lasso set-up. For the large dimensionality scenario, this would correspond to that a considerable proportion of the total candidate features being eliminated from the model. This can hardly happen in the ridge regression case, in which the shaded area is square rather than the rectangular, as the left panel of Fig. \ref{fig:L2L1Compare} illustrates.
\begin{algorithm}[h]
\caption{Coordinate descent algorithm for computing the Lasso}
\label{alg:Shooting}
\begin{algorithmic}[1]
\State Let $\bm{\beta}^{[0]} \in \mathbb{R}^{p}$ be the initial estimator. Set $s=0$.
\Repeat
    \State $s=s+1$
    \For{$j=1,\cdots,p$}
        \State $\beta_j^{[m]}=\frac{\operatorname{sign}(Z_j)(|Z_j|-\frac{\lambda}{2})_+}{\hat{\Sigma}_{jj}}$,
        \State  \multiline{%
        where $Z_j=\bm{X}_j^{\top}(\bm{Y}-\bm{X}\bm{\beta}_{-j}^{[m]})/n$, $\bm{\beta}_{-j}^{[m]}$ is the same as $\bm{\beta}^{[m]}$ except the $j$-th component is set to zero vector,  $\hat{\Sigma}=n^{-1}\bm{X}^{\top} \bm{X}$, and $\hat{\Sigma}_{jj}$ is the $j$-th diagonal component of $\hat{\Sigma}$}
    \EndFor
\Until numerical convergence
\end{algorithmic}
\end{algorithm}

In Algorithm \ref{alg:Shooting}, $(\cdot)_+ = \max(\cdot,0)$. The coordinate descent optimization deals with $\beta_1,\cdots, \beta_p$, whereas $\beta_0$ can be estimated by the empirical mean of the response variable. 

It is worth noting that the $L_1$ penalty on the coefficients in Lasso suggests that it only make sense if the different dimensions of features have the same ``strength''. Therefore it is the common practice of statisticians to standardize data before applying the Lasso (as well as for the ridge regression case). Namely, each covariate of data should be subtracted by the mean, and then be divided by the standard deviation. In this way, all the dimensions of data after the pre-processing procedure would have zero mean and unit variance.

The specification of the regularization parameter $\lambda$ plays a critical role. A large $\lambda$ will put more penalty on the coefficients, and shrink more weights to zero. In practice, one can select $\lambda$ that minimizes the cross-validation error. Another approach is to use the value of $\lambda$ that leads to one standard deviation of distance from the minimum cross-validation error. Both of these two approaches have been offered as an option in many softwares, e.g., Matlab. Comparing between these options, the first one leads to a slightly smaller prediction error, while the second approach usually render a more sparse model, i.e., a model composed of a noticeably smaller number of features. Nonetheless, whether a feature is useful in the model can only be examined by statistical testing. Therefore, the difference that these two approaches made in the final set of features does not mean that a group of features is useful according to the first approach but not useful according to the second. In our research, we adopt the first approach.  

\section{Results}\label{sec:Results}
\subsection{Forecasting the next-day's maximum $O_3$}\label{sbsec:Max}

First of all, we examine our approach by modeling the next-day's maximum $O_3$. After dealing with the missing data, we use the first three years' data (2014-2016) as the training data set, and use the year 2017's data as the testing dataset.

When we perform the Lasso described in Section \ref{sbsec:Lasso}, we use all the features in Table \ref{tab:Features}. Due to the design of the linear Lasso, we can see that unlike the multiple linear regression (MLR), adding a feature like $X_i-X_j$ makes a difference than merely having the feature $X_i$ and $X_j$, therefore we also include the terms representing the difference of the future day's meteorological variables and today's value of the same hour (as well as the next-day's max/min/mean minus the current day's max/min/mean). Since we have 7 different pollutants available, there are $7\times 24 +7\times 3=189$ non-meteorological features. Besides, when we are dealing with the wind direction within the meteorological variables, we not only include its value in degrees, but also in cosine and sine values. Since we have 7 different kinds of meteorological variables available, and we have $(7+2)\times (24+3) \times 3 = 729$ meteorological features. Here $7+2$ corresponds the original 7 features plus the cosine and sine of the wind direction, $24+3$ corresponds to the 24 hours values plus the max/min/mean, and the last multiplier $3$ indicates the values of the current day, the future day, and their differences. In total, there are $189+729=918$ features in our original model.

Also, rather than using the future-day's maximum concentration value of $O_3$ as the model output, we use the value of the future-day's maximum minus the current day's maximum as the target for the model output. As discussed in Section \ref{sbsec:Lasso}, all the variables are standardized before the modeling, and the ``shooting algorithm'' as in Algorithm 1 is used to solve the Lasso. In order to select the regularization parameter $\lambda$ as in (\ref{eq:LassoCr}), we use the 5-fold cross validation and select the value that minimizes the cross-validation error. In this way, $\lambda$ is chosen as $0.0121$. 

After the modeling solved by the Lasso, we find that among the $918$ original features, only $120$ of them correspond to nonzero weight in the final model. In other words, 798 of the original 918 features have been eliminated from the model. Then we use the testing dataset to evaluate the prediction accuracy, and we find that the root mean square error (RMSE) to be 6.12 ppb, and the mean absolute error (MAE) to be 4.92 ppb. 

Further from the sparse modeling of the aforementioned 918 linear features solved by the Lasso, we can also include the interactive features expanded by the product of those 918 features. In this way, the modeling is conducted onto a $918+918\times917/2+918= 422739$ feature space. Among them, there are terms like the original variables (918 of them) and their second polynomial terms (918 of them), and there are terms like wind speed multiplied by cosine of wind direction ($918\times917/2$ of them). The regularization parameter $\lambda$ found by a 5-fold cross validation is set to be 0.0295, and we find only 193 features of the total 422739 candidate features exist in the final model. After applying the model to the testing dataset, we find the RMSE= 5.63 ppb, and MAE= 4.42 ppb.

It is worth mentioning that if we just use the current day's maximum value of $O_3$ concentration to approximate the next day's maximum $O_3$ concentration, then the RMSE=9.21 ppb, and MAE=6.81 ppb.

In order to compare our modeling approach with other machine learning methods, we compare the modeling using the multiple linear regression (MLR) \cite{barrero2006prediction,ghazali2010transformation,sousa2006prediction}, and the ridge regression \cite{salazar2008development} which also were described in Section \ref{sbsec:Lasso}. Both methods are conducted on the same 918 feature space. Apart from that, we also compare with the time-series modeling \cite{robeson1990evaluation}, in which we consider the daily maximum value as a ARMA process. Thirdly, we also compared with the Support Vector Regression (SVM-regression) approach \cite{hajek2012ozone,lu2008ground}. The prediction accuracy of these techniques are shown in Table \ref{tab:ResultMax}.

\begin{table}[width=.9\linewidth,cols=4,pos=h]
\caption{The prediction of daily maximum $O_3$ concentration using various statistical and machine learning methods. The RMSE and MAE values are in the unit of ppb. The last column of this table shows the final number of features in the model, as well as the number of candidate features originally examined by the model. }\label{tab:ResultMax}
\begin{tabular*}{\tblwidth}{@{} LLLL@{} }
\toprule
~ & RMSE & MAE  & \#Features\\
\midrule
Lasso (linear)    & 6.12 & 4.92 & 105/ 918 \\
\midrule
Lasso (polynomial)& 5.63 & 4.42 & 193/ 422739 \\
\midrule
ridge regression  & 8.15 & 6.52 & 918/ 918 \\
\midrule
ARMA (time series)& 9.26 & 6.86 & n/a \\
\midrule
MLR               & 11.52& 8.39 & 918/ 918 \\
\midrule
SVM regression    & 9.32 & 7.53 & 918/ 918 \\
\midrule
\scriptsize{Use previous day's value} & 9.21 & 6.81 & n/a \\
\bottomrule
\end{tabular*}
\end{table} 

From Table \ref{tab:ResultMax}, we can see that our Lasso approach on the original 918 candidate features already outperforms the other competing methods, moreover, conducting the Lasso on the interactive features (polynomial model) yields even stronger prediction accuracy.

\begin{figure}
	\centering
	\subfloat[]{\includegraphics[width=85mm]{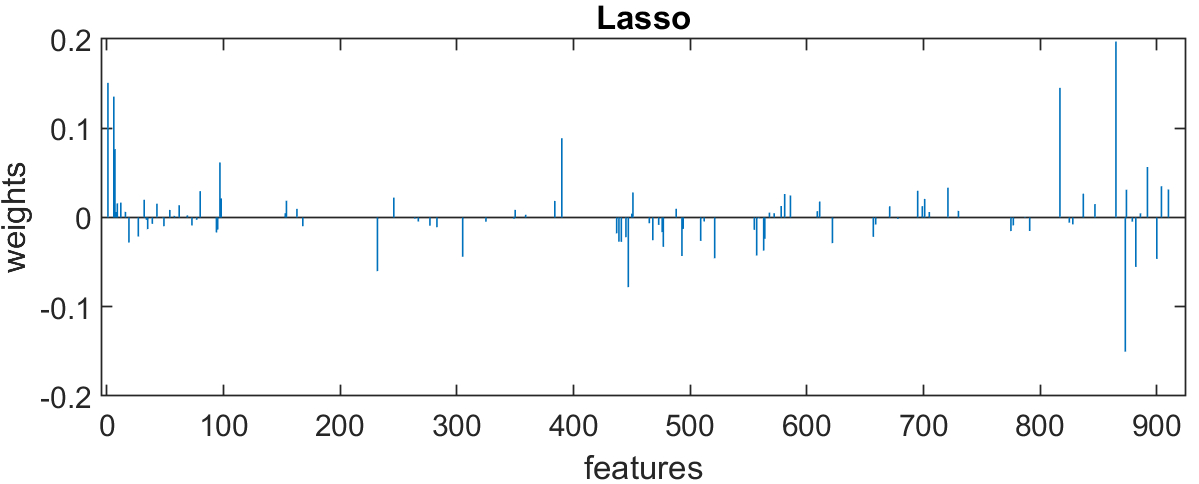}}\\
	\subfloat[]{\includegraphics[width=85mm]{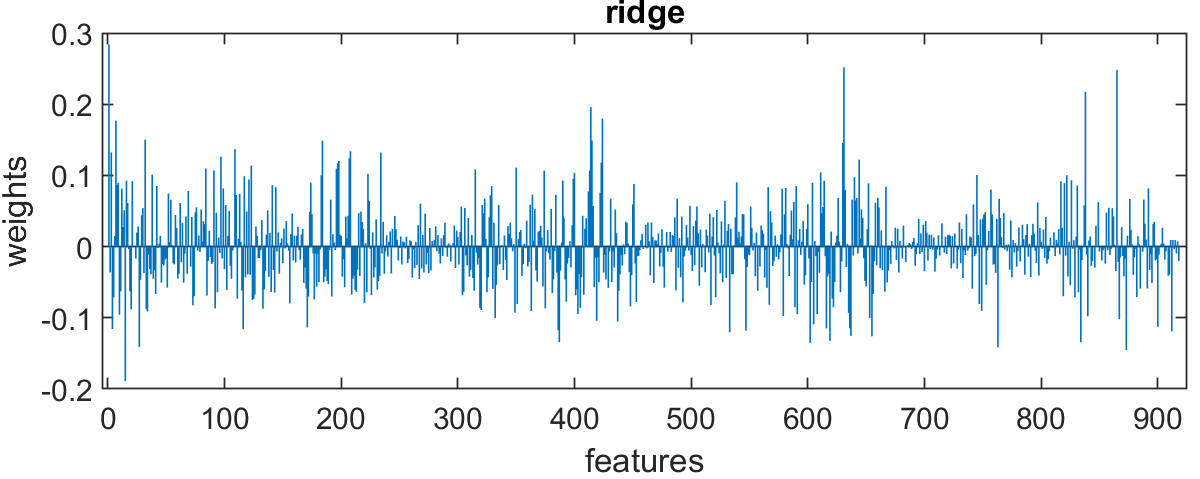}}
	\caption{The weights in the final forecast model for the next-day's maximum $O_3$ obtained through
		(a) Lasso method (b) ridge regression.} \label{fig:MaxWeights}
\end{figure}

The model weight $\beta_j$ corresponds to the linear Lasso are shown in Fig. \ref{fig:MaxWeights} (a). Since the features have been standardized, therefore the strength of each feature can be compared to each other, i.e., the features corresponds to the highest value of weight is the most important one in the model. From Fig. \ref{fig:MaxWeights} (a), we can clearly see that the majority of features have been eliminated. On the other hand, the weights correspond to the ridge regression are shown in Fig. \ref{fig:MaxWeights} (b). The comparison between these two graphs clearly demonstrate the difference between the nature of $L_1$ and $L_2$ design leads to the unique automatic feature selection property inherent to the Lasso approach, which explains the superiority of the Lasso's prediction accuracy.

In accordance to Fig. \ref{fig:MaxWeights} (a), if we show the ten (10) most important features and their strength, we obtain Table \ref{tab:10FMax}. It is worth to mention that these features here have already been standardized in the pre-processing. 

\begin{table}[width=.9\linewidth,cols=4,pos=h]
\caption{The first 10 dominant features in the linear model solved by Lasso to predict the next-day's maximum $O_3$ concentration.}\label{tab:10FMax}
\begin{tabular*}{\tblwidth}{@{} LL@{} }
\toprule
Weight & Feature \\
\midrule
 0.1971 & \scriptsize{next-day's maximum temperature}   \\
\midrule
 0.1508 & \scriptsize{current-day's $O_3$ concentration at 11 pm}    \\
\midrule
 -0.1502 & \scriptsize{next-day's minimum relative humidity}    \\
\midrule
 0.1453 & \scriptsize{current-day's maximum $O_3$ concentration}    \\
\midrule
 0.1355 & \scriptsize{current-day's $O_3$ concentration at 6 pm}    \\
\midrule
 0.0888 & \scriptsize{next-day's temperature at 6 am}    \\
\midrule
 -0.0780 & \scriptsize{next-day's relative humility at 3 pm}    \\
\midrule
 -0.0766 & \scriptsize{current-day's $O_3$ concentration at 5 pm}    \\
\midrule
0.0617 & \scriptsize{current-day's $NO_2$ concentration at 11 pm}    \\
\midrule
-0.0601 & \scriptsize{current-day's cosine of wind direction at 8 am}    \\
\midrule
\vdots  & \vdots \\
\bottomrule
\end{tabular*}
\end{table} 


\begin{figure}
	\centering
	\subfloat[]{\includegraphics[width=85mm]{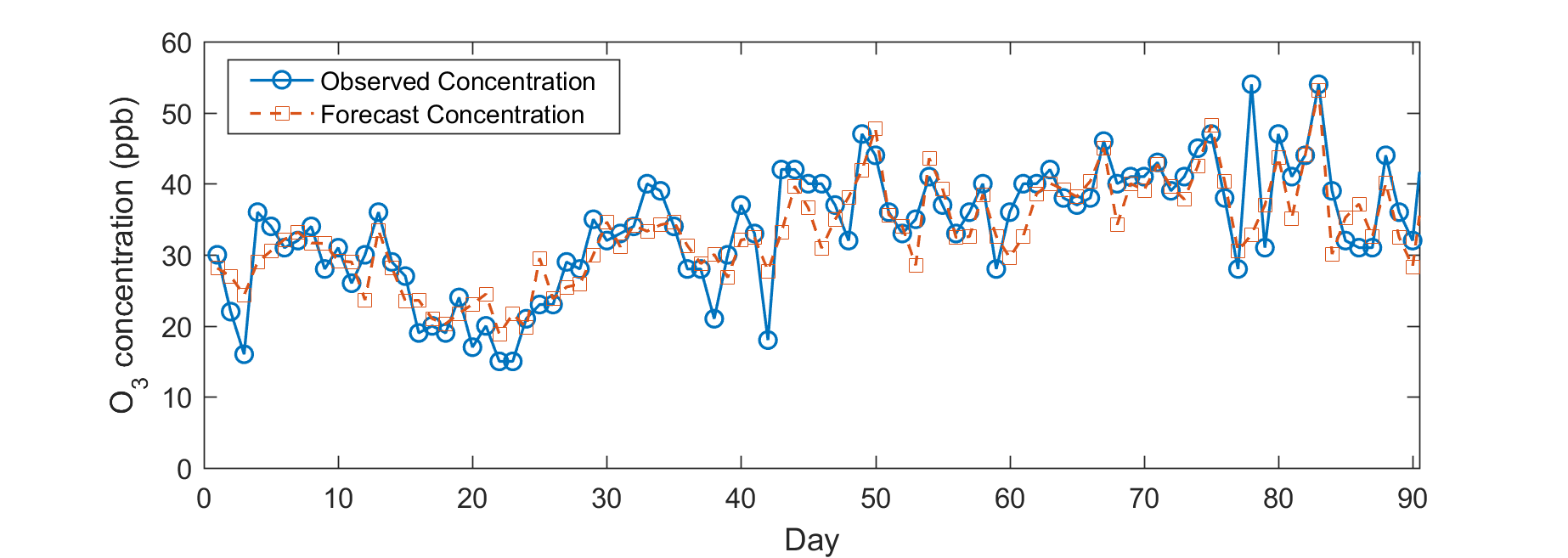}}\\
	\subfloat[]{\includegraphics[width=85mm]{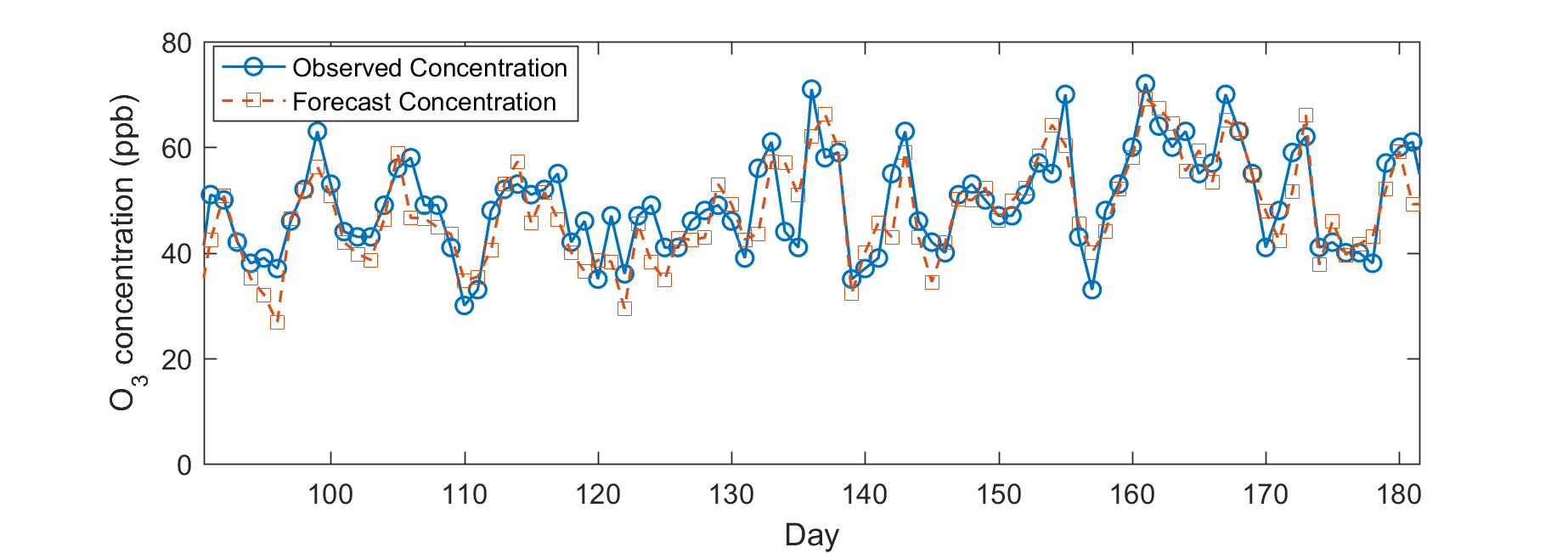}}\\
	\subfloat[]{\includegraphics[width=85mm]{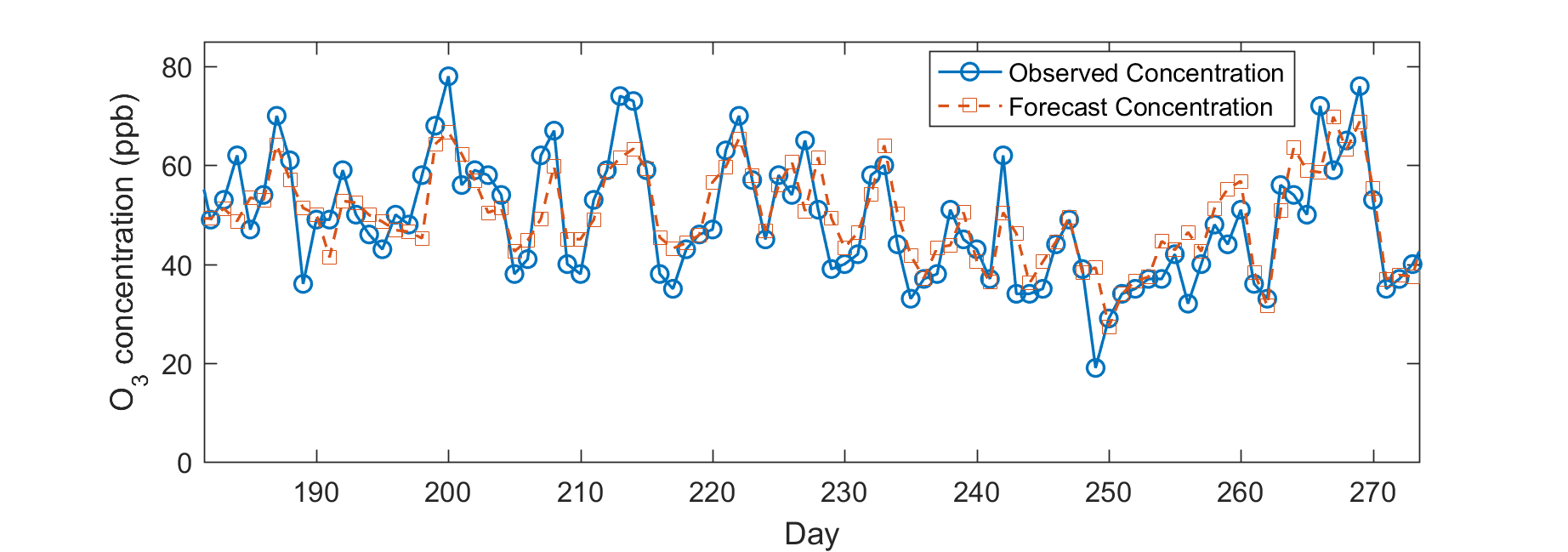}}\\
	\subfloat[]{\includegraphics[width=85mm]{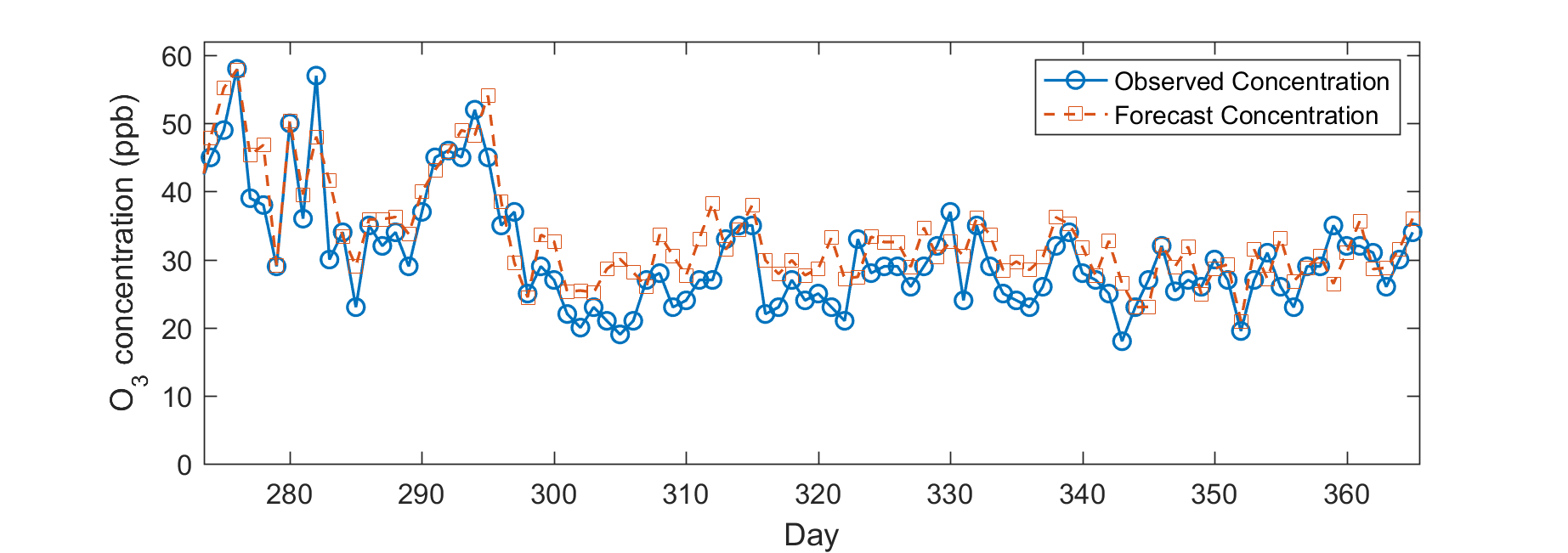}}
	\caption{The prediction of the next-day's maximum $O_3$ concentration using the polynomial model solved by the Lasso. (a) The 1st trimester, (b) the 2nd trimester, (c) the 3rd trimester, (d) the 4rd trimester.} \label{fig:MaxPredict}
\end{figure}

We also plot the actual values of the daily maximum $O_3$ concentration in the test dataset together with the predicted values using our modeling with the linear as well as the interactive features (the modeling using the 422739 candidate features). The result is shown in Fig. \ref{fig:MaxPredict}.

\begin{figure}
	\centering
	\subfloat[]{\includegraphics[width=43mm]{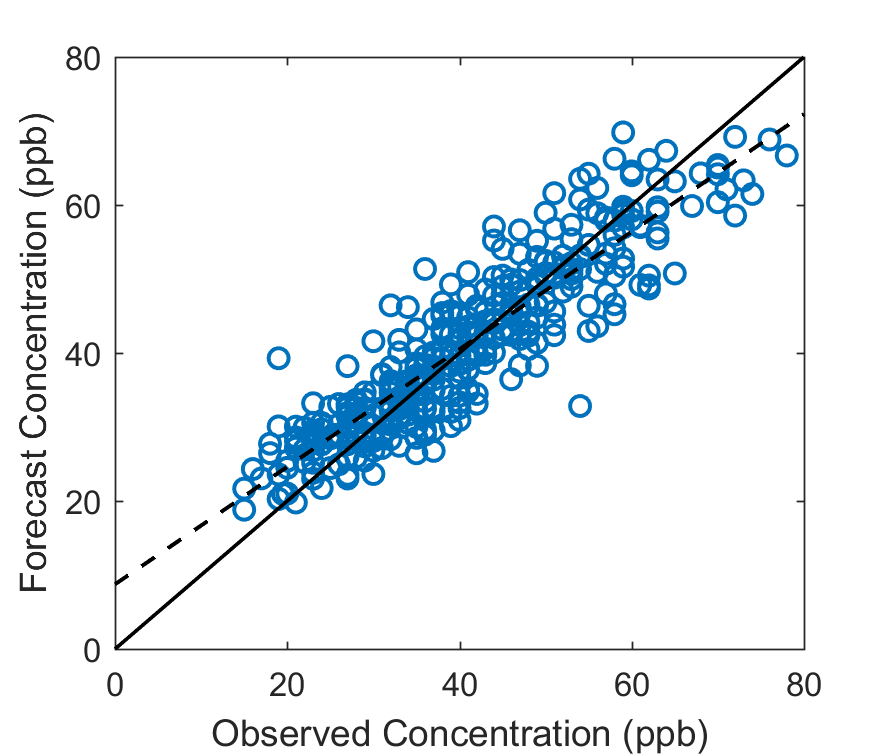}}\\
	\subfloat[]{\includegraphics[width=43mm]{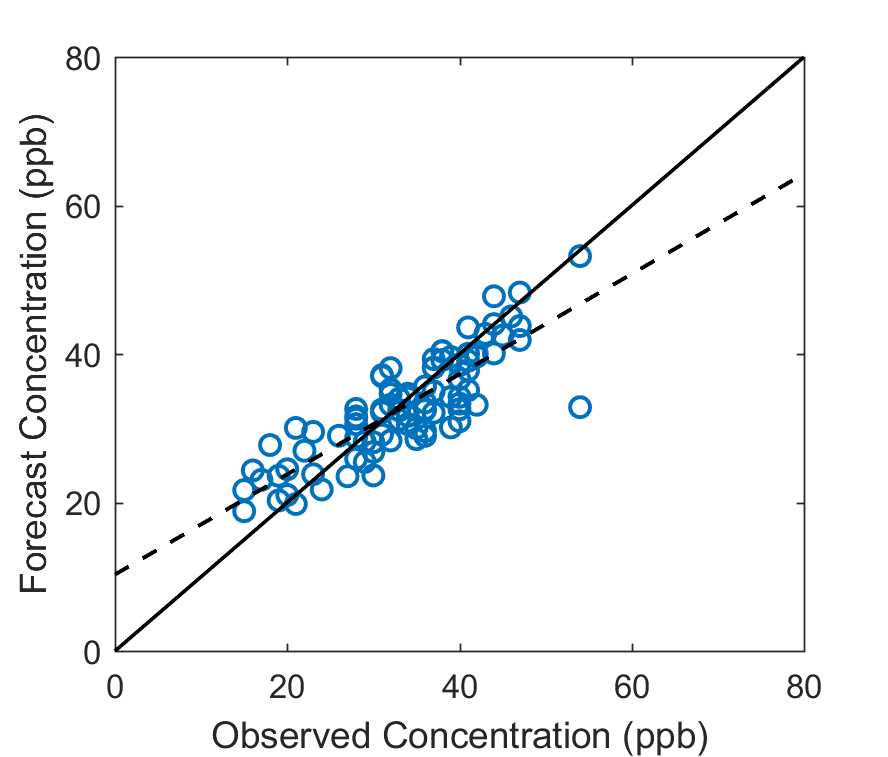}} ~\
	\subfloat[]{\includegraphics[width=43mm]{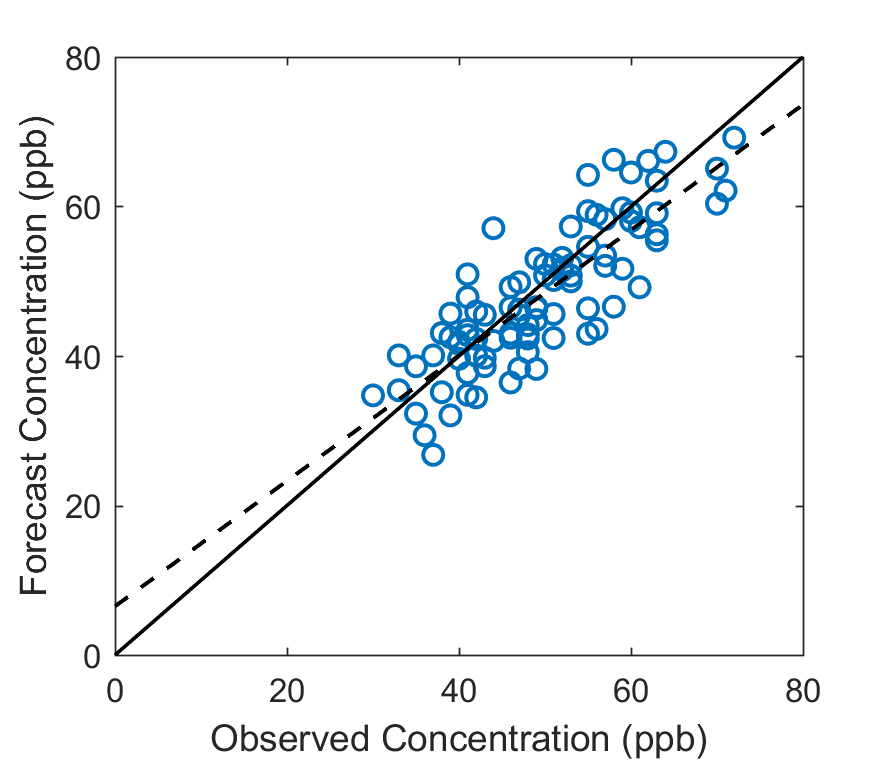}}\\
	\subfloat[]{\includegraphics[width=43mm]{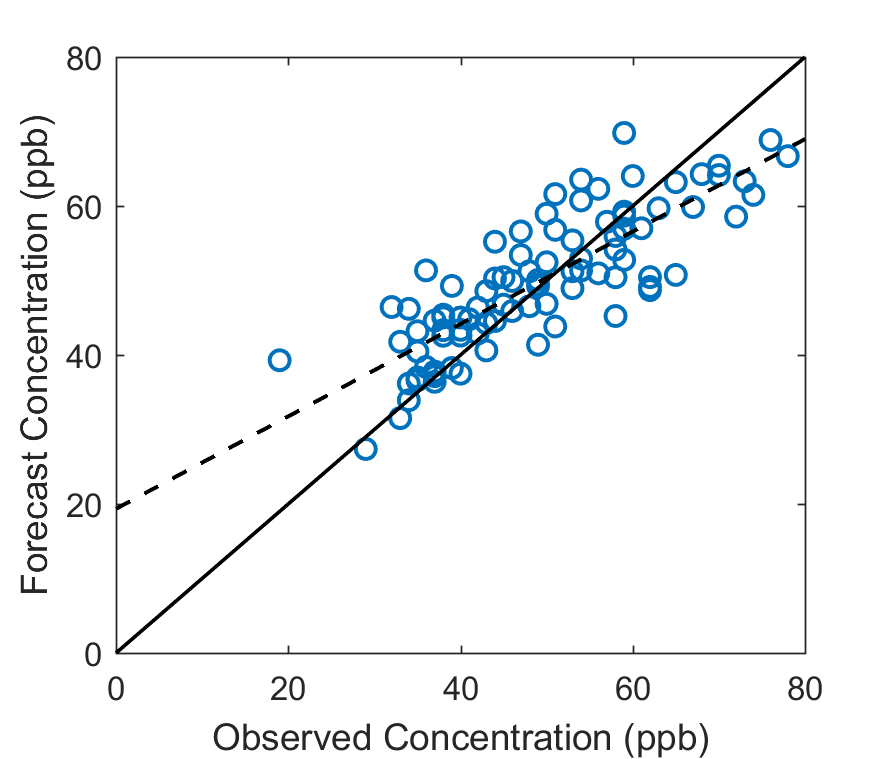}} ~\
	\subfloat[]{\includegraphics[width=43mm]{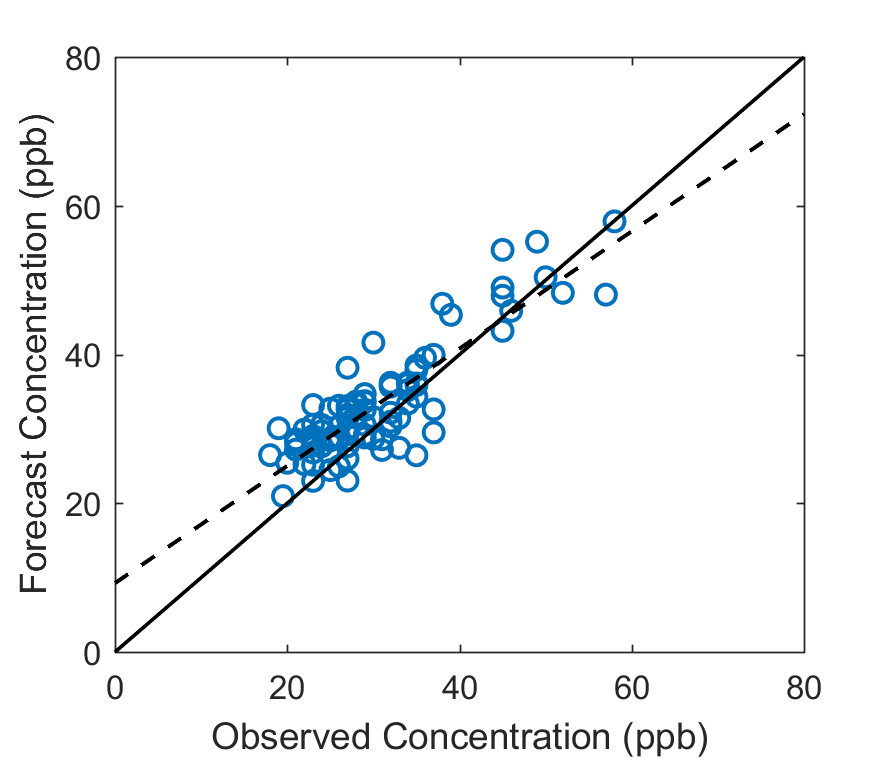}}
	\caption{Scatterplots of forecast vs observed next-day maximum ozone concentrations for the testing set. (a) The whole year, (b) the 1st trimester, (c) the 2nd trimester, (d) the 3rd trimester, (e) the 4th trimester. Note that the solid line represents the perfect prediction while the dashed line  is an ordinary least square fit in each graph.} \label{fig:MaxScatter}
\end{figure}

If we examine the scatterplot of the predicted concentration value vs the true value, we obtain Fig. \ref{fig:MaxScatter}. We plot not only the whole year but also each trimester. From Fig. \ref{fig:MaxScatter}, we find that our modeling predict the best for the second trimester among all the four.

\subsection{Forecasting the next-day's maximum of 8-hour-mean $O_3$}\label{sbsec:8HourMeanMax}

The maximum of the 8-hour-mean value has played an important role in environmental monitor and control \cite{link3}. The definition of this concept is that the mathematical average of a consecutive eight hours' ozone concentration values is recorded to be the ``8-hour-mean'' value at the first hour. The daily maximum of this 8-hour-mean concentration is not necessarily the highest concentration of this variable in a whole day. Similar to Section \ref{sbsec:Max}, three whole years data (2014-2016) are used as the training dataset, and the year 2017's data are used as the testing dataset.

\begin{figure}
	\centering
	\subfloat[]{\includegraphics[width=85mm]{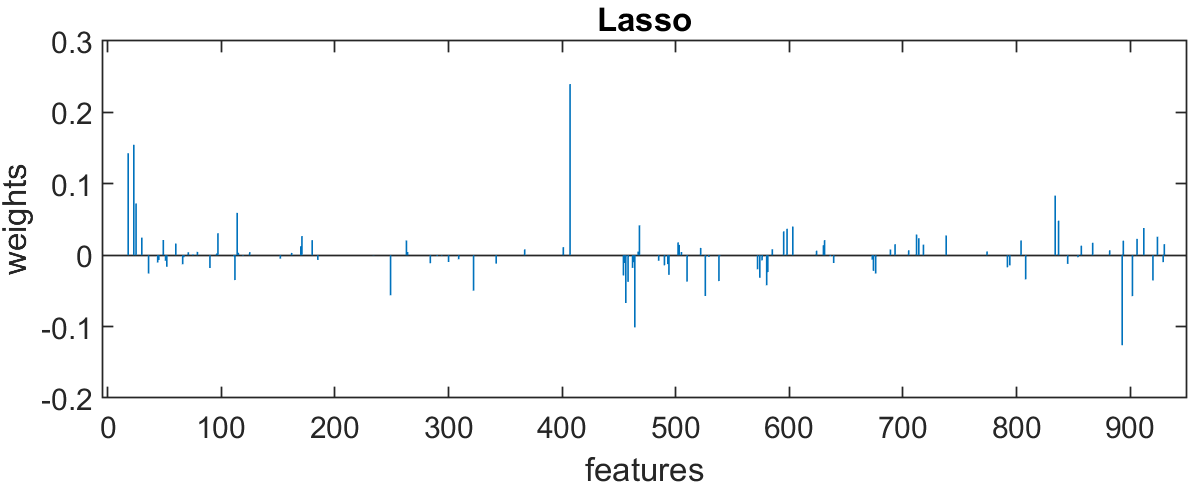}}\\
	\subfloat[]{\includegraphics[width=85mm]{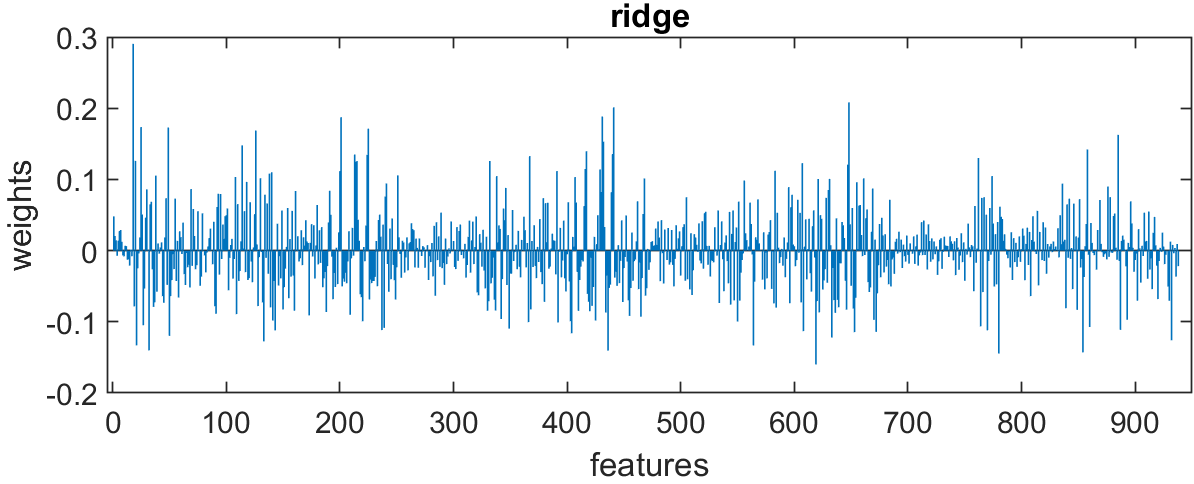}}
	\caption{The weights in the final forecast model for the next-day's maximum 8-hour-mean $O_3$ obtained through
		(a) Lasso method, (b) ridge regression.} \label{fig:Max8HMWeights}
\end{figure}

When the Lasso is performed for the linear model, all the same 918 features as described in Section \ref{sbsec:Max} are used. Apart from that, we also use the 8-hour-mean values of $O_3$ concentration at 0 am - 4 pm of the current-day. It is worth mentioning in the current day, we only have the 8-hour-mean value at up to 4 pm, which actually is composed of the hourly value at 11 pm. Together with the maximum, mean, minimum of those seventeen (17) 8-hour-mean values, the total numbers of candidate features in the linear model is thus 938. A 5-fold cross validation is applied to select the regularization parameter $\lambda$, and we use $\lambda=0.0118$. By using the Lasso, the number of final features in the model is reduced to 113. In Fig. \ref{fig:Max8HMWeights}, the weights selected by Lasso is compared to the weights selected by ridge regression, and in Table \ref{tab:ResultMax8HM}, it shows the prediction performance of various techniques.

The second row of Table \ref{tab:ResultMax8HM} shows the performance of the Lasso with the interactive features (polynomial model). Similar to the way in Section \ref{sbsec:Max}, the 938 linear features are expanded into $938+938 \times 937/2+938=441329$ features. 

In Table \ref{tab:ResultMax8HM}, our proposed modeling approach is compared with the other competing methods mentioned in Section \ref{sbsec:Max}.Table \ref{tab:ResultMax8HM} again confirms the two Lasso approaches (linear model and polynomial model) outperform the other ones.  

\begin{table}[width=.9\linewidth,cols=4,pos=h]
\caption{The prediction of daily maximum 8-hour-mean $O_3$ concentration using various statistical and machine learning methods. The RMSE and MAE values are in the unit of ppb. The last column of this table shows the final number of features in the model, as well as the number of candidate features originally examined by the model.}\label{tab:ResultMax8HM}
\begin{tabular*}{\tblwidth}{@{} LLLL@{} }
\toprule
~ & RMSE & MAE  & \#Features\\
\midrule
Lasso (linear)    & 6.20 & 4.85 & 113/ 938 \\
\midrule
Lasso (polynomial)& 5.68 & 4.52 & 160/ 441329 \\
\midrule
ridge regression  & 7.87 & 6.18 & 938/ 938 \\
\midrule
ARMA (time series)& 10.51 & 7.65 & n/a \\
\midrule
MLR               & 11.14& 8.02 & 938/ 938 \\
\midrule
SVM regression    & 9.06 & 7.16 & 938/ 938 \\
\midrule
\scriptsize{Use previous day's value} & 8.98 & 6.52 & n/a \\
\bottomrule
\end{tabular*}

\end{table}

The comparison between the weights selected by Lasso and ridge regression in the linear set-up is shown in Fig. \ref{fig:Max8HMWeights}. Again, this shows the nature of $L_1$ design allows the optimal feature selection, which leads to better prediction accuracy. The ten (10) most prominent features in the resultant Lasso linear model is shown in Table \ref{tab:10FMax8HM}. Again, these features here have already been standardized in the pre-processing.

\begin{table}[width=.9\linewidth,cols=4,pos=h]
\caption{The first 10 dominant features in the linear model solved by Lasso to predict the next-day's maximum 8-hour-mean $O_3$ concentration.}\label{tab:10FMax8HM}
\begin{tabular*}{\tblwidth}{@{} LL@{} }
\toprule
Weight & Feature \\
\midrule
 0.2395 & \scriptsize{next-day's temperature at 6 am}   \\
\midrule
 0.1546 & \scriptsize{current-day's $O_3$ concentration at 6 pm}    \\
\midrule
 0.1429 & \scriptsize{current-day's $O_3$ concentration at 11 pm}    \\
\midrule
 -0.1261 & \scriptsize{next-day's minimum relative humility}    \\
\midrule
 -0.1011 & \scriptsize{next-day's relative humility at 3 pm}    \\
\midrule
 0.0835 & \scriptsize{current-day's maximum 8-hour-mean $O_3$ concentration}    \\
\midrule
 0.0724 & \scriptsize{current-day's $O_3$ concentration at 4 pm}    \\
\midrule
 -0.0670 & \scriptsize{next-day's atmospheric pressure at 7 am}    \\
\midrule
0.0592 & \scriptsize{current-day's $NO_2$ concentration at 11 pm}    \\
\midrule
-0.0574 & \scriptsize{next-day's minimum visibility}   \\
\midrule
\vdots  & \vdots \\
\bottomrule
\end{tabular*}
\end{table} 


It is worth mentioning that the prediction accuracy of the daily maximum 8-hour-mean $O_3$ concentration is a little worse than that of the daily maximum concentration in Section \ref{sbsec:Max}. The reason could be that the next-day's maximum 8-hour-mean $O_3$ concentration could sometimes happen after 4 pm, which include the information in the day after the next day.  On the other hand, comparing Table \ref{tab:10FMax} and Table \ref{tab:10FMax8HM}, we can see that the same seven (7) features are present in both tables. (We consider the maximum 8-hour-mean $O_3$ concentration in Table \ref{tab:10FMax8HM} to be the counterpart of maximum $O_3$ concentration in Table \ref{tab:10FMax}.) However, the sequence and weights of these 7 identical features are different, and the most dominating feature in the two tables are different. The reason is probably due to the fact that the information in many important pollutants/meteorological variables are partially included in each other in a profoundly interactive way. Among the various pollutants, it seems the concentration of $NO_2$ can greatly contribute to predict the $O_3$ concentration accurately. Among the meteorological variables, the temperature and the relative humility play an significant role. Regarding to some other pollutants and meteorological variables, their influence seems to be less significant, but the elimination of their influence can only be concluded through statistical testings in a rigorous way, e.g., the lack-of-fit model tests, which is beyond the scope of this paper.

\begin{figure}
	\centering
	\subfloat[]{\includegraphics[width=85mm]{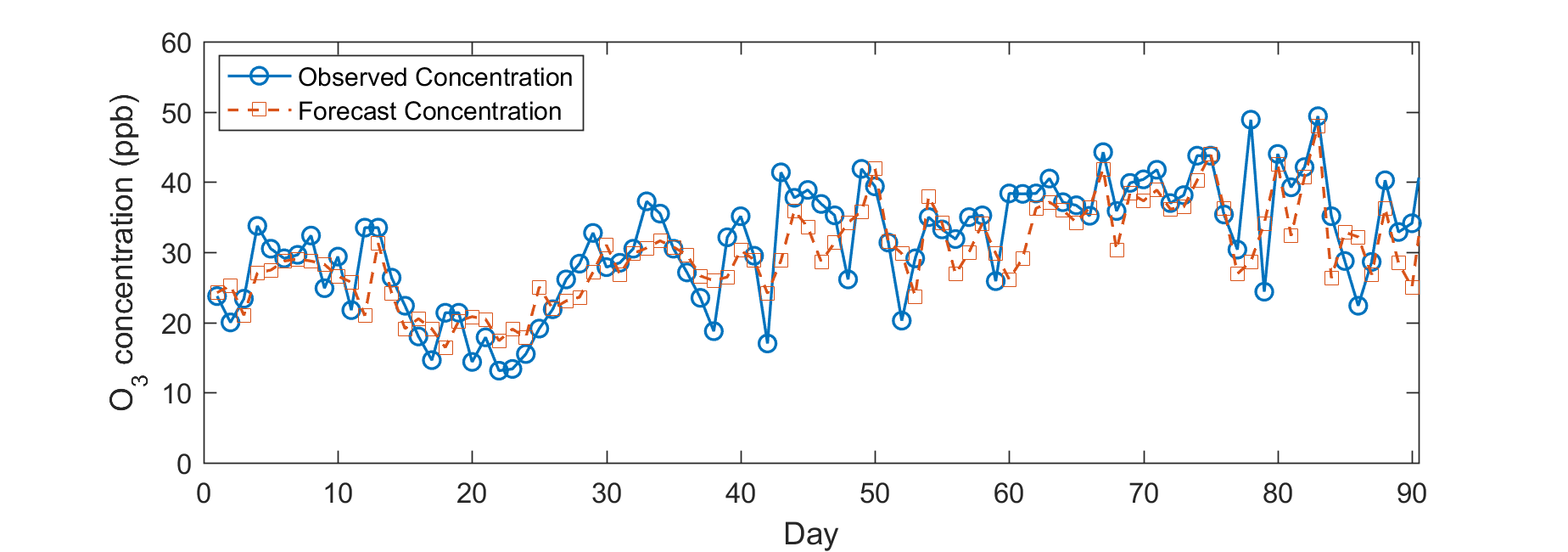}}\\
	\subfloat[]{\includegraphics[width=85mm]{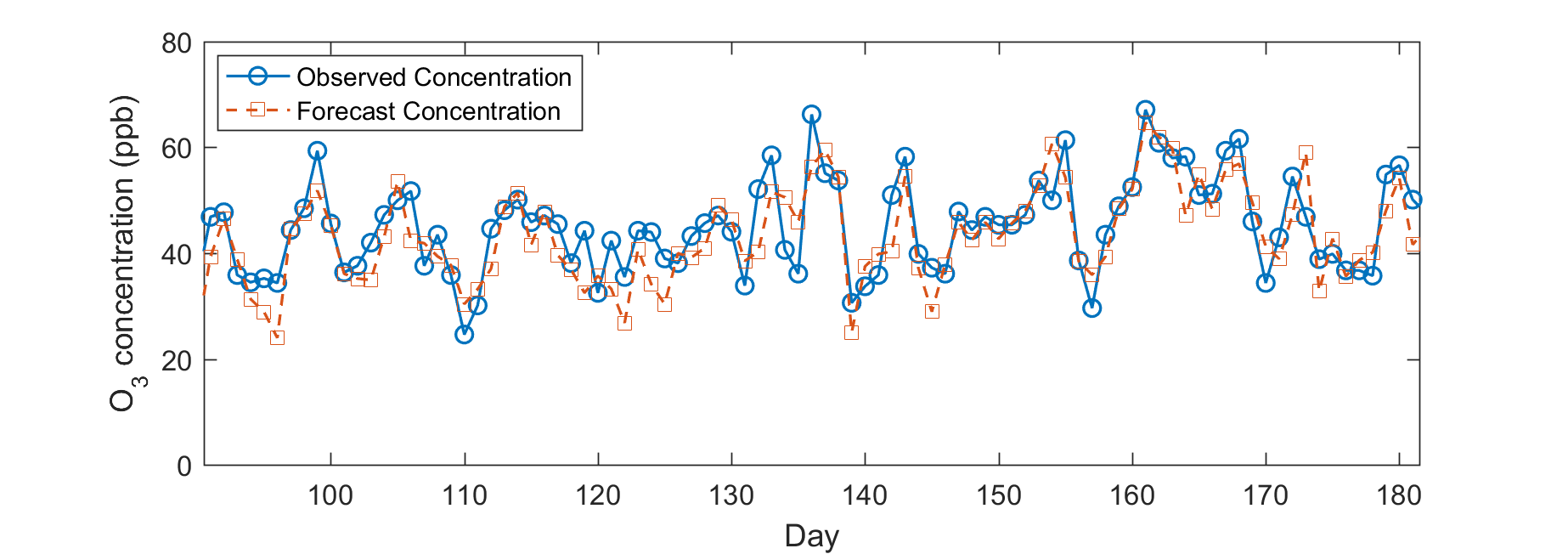}}\\
	\subfloat[]{\includegraphics[width=85mm]{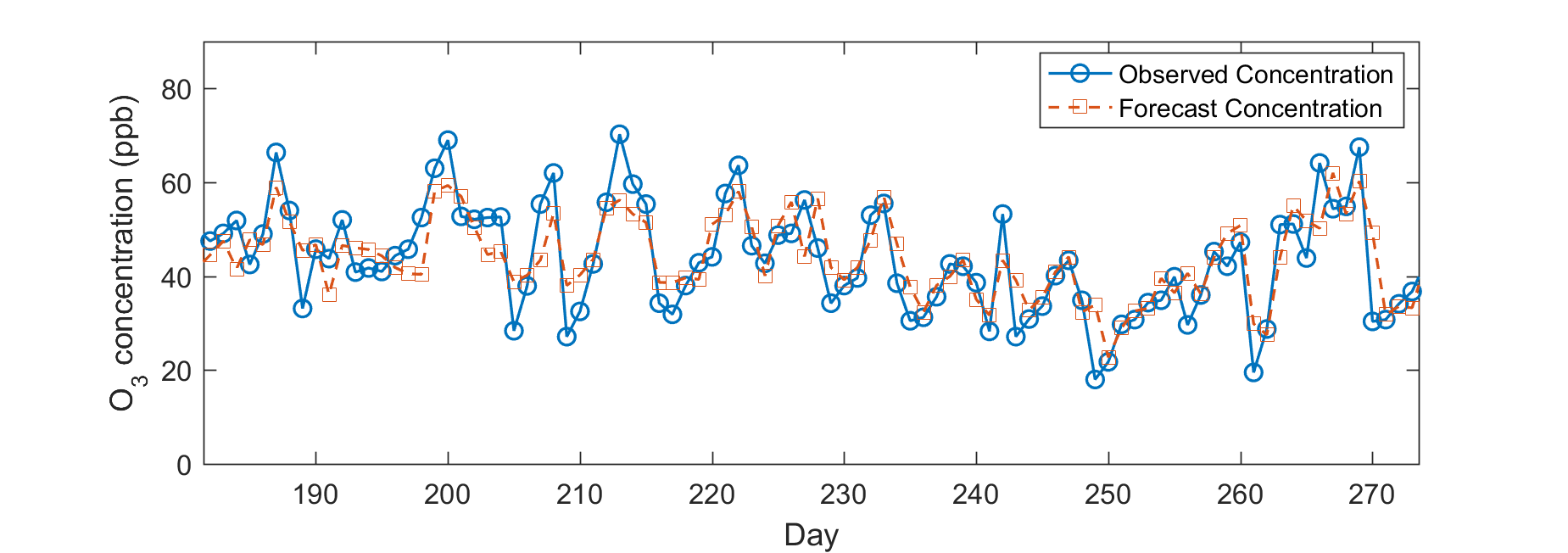}}\\
	\subfloat[]{\includegraphics[width=85mm]{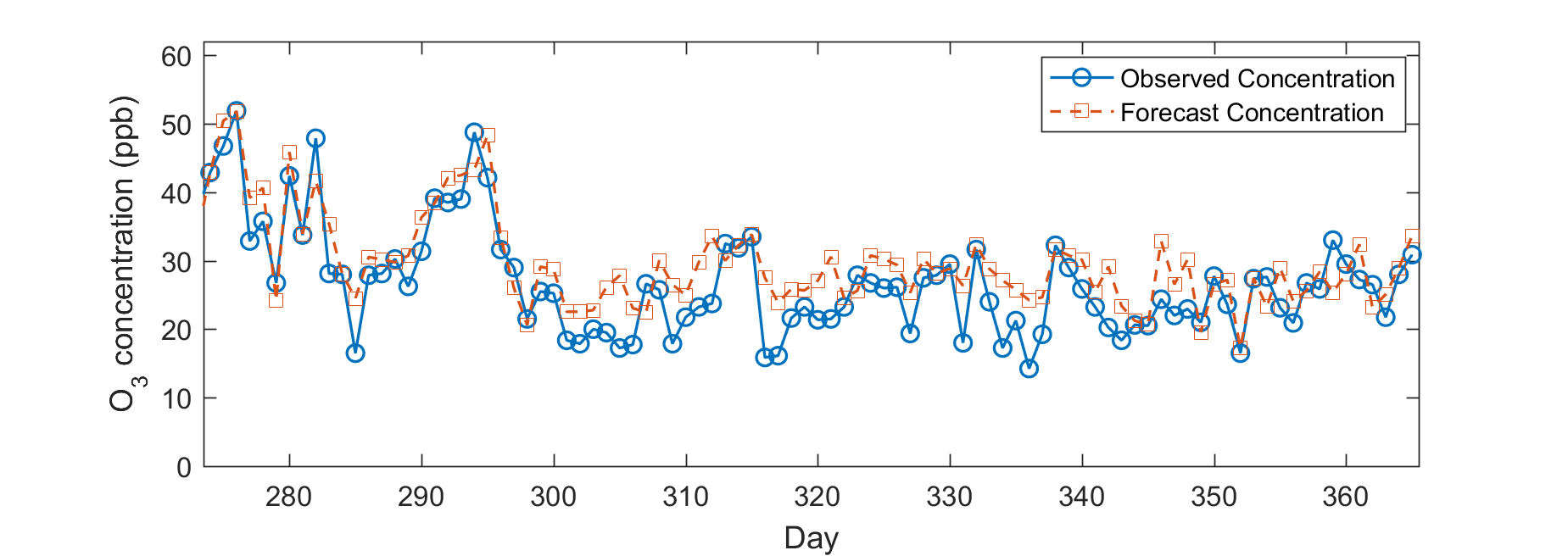}}
	\caption{The prediction of the next-day's maximum 8-hour-mean $O_3$ concentration using the polynomial model solved by the Lasso. (a) The 1st trimester, (b) the 2nd trimester, (c) the 3rd trimester, (d) the 4rd trimester.} \label{fig:Max8HMPredict}
\end{figure}

For the performance of Lasso polynomial model (interactive features), the true daily maximum 8-hour-mean $O_3$ concentration values in the testing set as well as the predicted values are shown in Fig. \ref{fig:Max8HMPredict}.

\begin{figure}
	\centering
	\subfloat[]{\includegraphics[width=43mm]{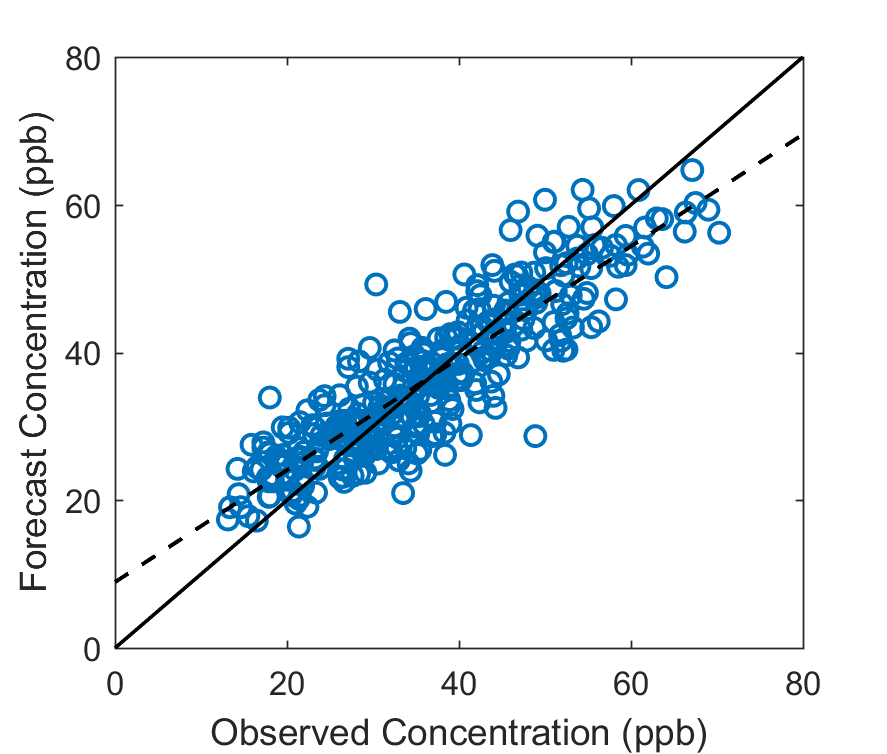}}\\
	\subfloat[]{\includegraphics[width=43mm]{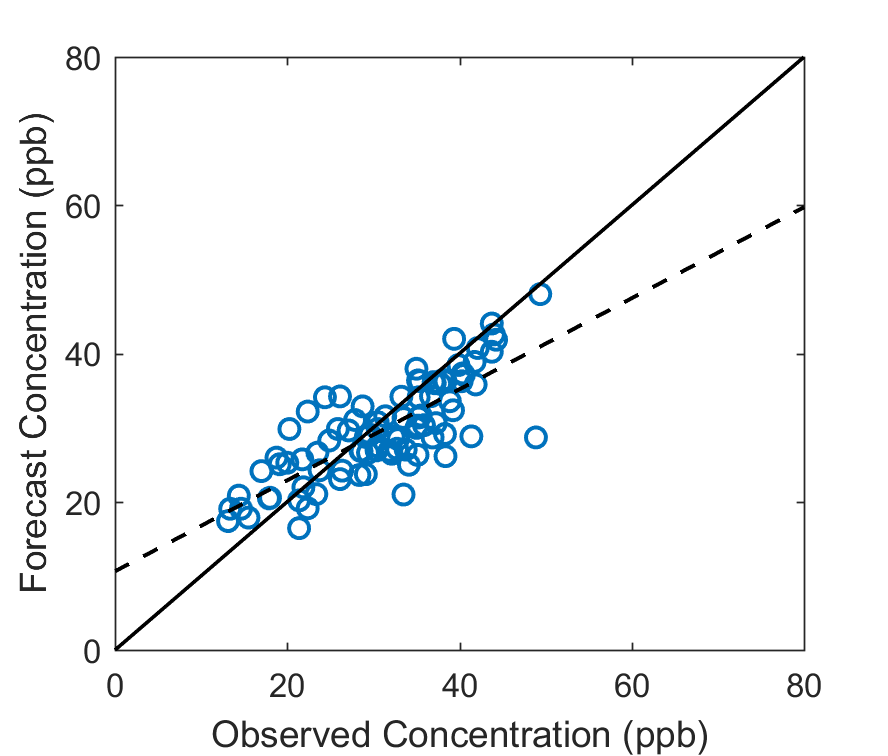}} ~\
	\subfloat[]{\includegraphics[width=43mm]{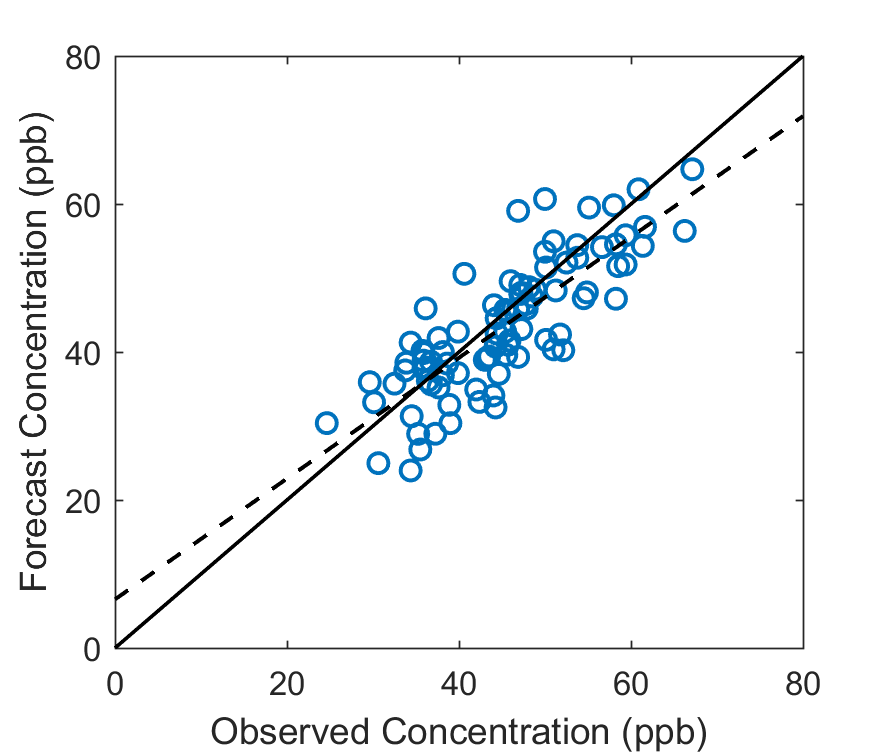}}\\
	\subfloat[]{\includegraphics[width=43mm]{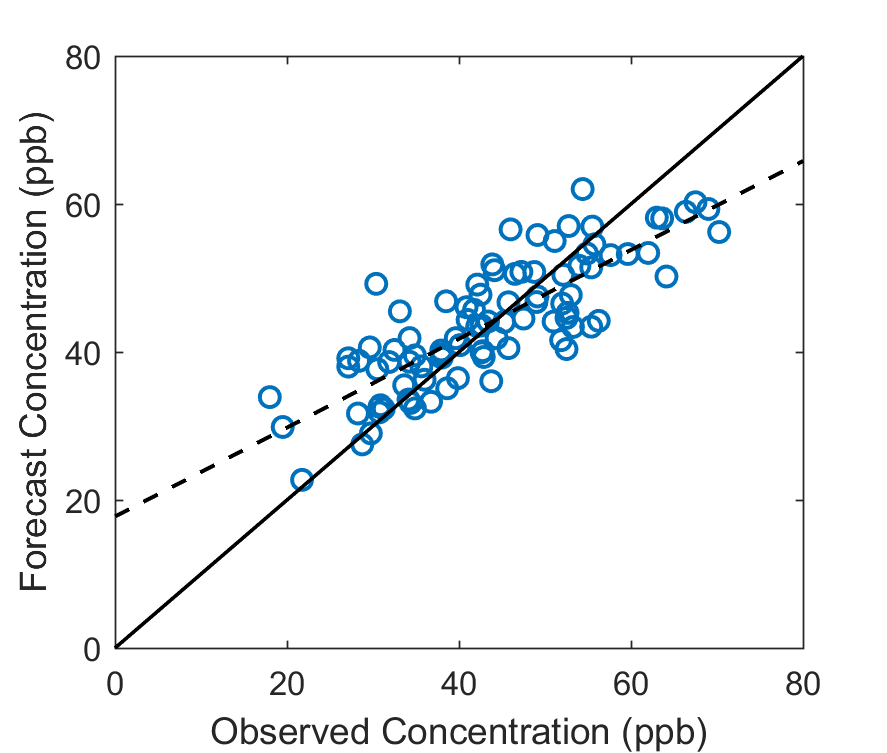}} ~\
	\subfloat[]{\includegraphics[width=43mm]{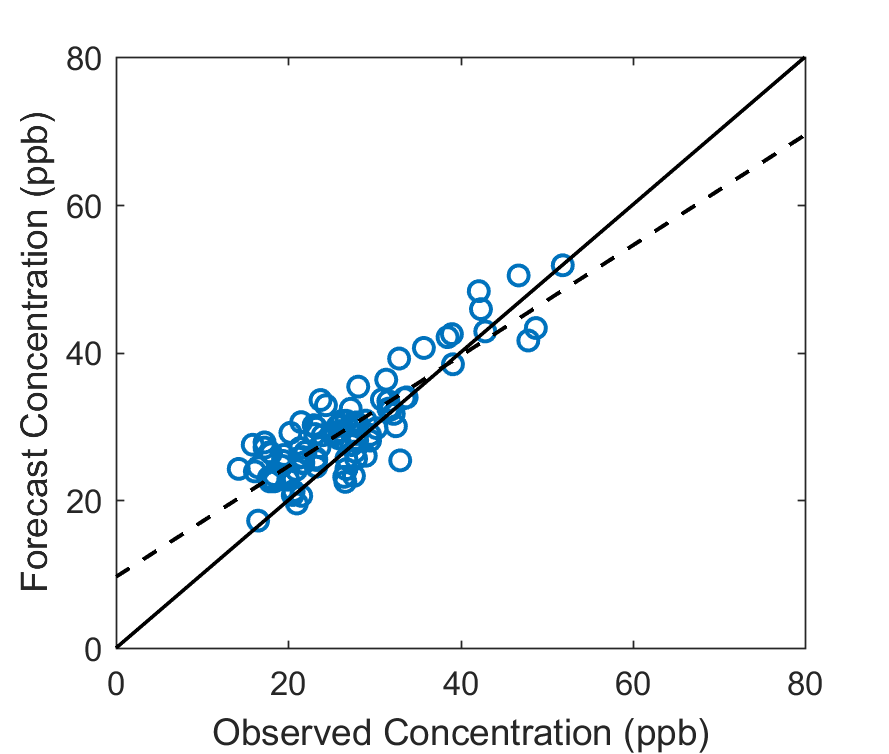}}
	\caption{Scatterplots of forecast vs observed next-day maximum 8-hour-mean ozone concentrations for the testing set. (a) The whole year, (b) the 1st trimester, (c) the 2nd trimester, (d) the 3rd trimester, (e) the 4th trimester. Note that the solid line represents the perfect prediction while the dashed line  is an ordinary least square fit in each graph.} \label{fig:Max8HScatter}
\end{figure}

If we examine the scatterplot of the predicted concentration value vs the true value for the maximum eight-hour-mean concentration value, we obtain Fig. \ref{fig:Max8HScatter}. We plot not only the whole year but also each trimester. Similar to the maximum daily concentration forecast, we find that our modeling predict the best for the second trimester among all the four.

\section{Discussions and Conclusion}\label{sec:Discussions}

In this paper, the Lasso approach is used for prediction of the $O_3$ concentration in terms of the next-day's maximum value, as well as the next-day's maximum 8-hour-mean value. The simulation studies show that this approach outperforms some other competing methods recently applied in the field.

It is worth mentioning that rather than modeling directly the next-day's maximum concentration (or next-day's maximum 8-hour-mean value), we also have tried the approach to target the difference between the next-day's and current-day's values. We achieve similar level of prediction accuracy. Therefore it is conjured that in case that the Lasso is applied in cities where the $O_3$ concentrations are usually significantly larger than the levels in Windsor, Canada, it will likely render prediction accuracy similarly in RMSE (or MAE) as the level we achieved in this paper.

\bibliographystyle{cas-model2-names}

\bibliography{ozoneRef,forecastLZ}


%
%

\end{document}